\documentclass[11pt,a4paper]{article}
\usepackage[hyperref]{acl2021}
\usepackage{times}
\usepackage{latexsym}
\usepackage{graphicx}
\usepackage{amsmath}
\usepackage{amsfonts}
\usepackage{booktabs}
\usepackage{multirow}
\usepackage{float}
\usepackage{times}
\usepackage{algorithm}
\usepackage{algorithmic}
\usepackage{enumitem}
\usepackage{amsfonts}
\usepackage{dsfont}
\usepackage{amssymb}
\usepackage{amsfonts}
\usepackage{bm}
\usepackage{bbm}
\usepackage{xspace}
\usepackage{helvet}
\usepackage{linguex}
\usepackage{xcolor}
\usepackage{cite}
\newcommand{\mname}{\textsc{ERNIE-Doc}\xspace}

\newcommand\blfootnote[1]{%
\begingroup
\renewcommand\thefootnote{}\footnote{#1}%
\addtocounter{footnote}{-1}%
\endgroup
}
\usepackage{microtype}

\aclfinalcopy 

\author{
    Siyu Ding$^*$,~~Junyuan Shang$^*$,~~Shuohuan Wang,~~Yu Sun,~~Hao Tian, \\ 
    \textbf{Hua Wu} and \textbf{Haifeng Wang} \\
    Baidu Inc., China\\
    \{\texttt{dingsiyu, shangjunyuan, wangshuohuan, sunyu02, } \\
    \texttt{tianhao, wu\_hua, wanghaifeng\}@baidu.com}
}
  
\title{\textsc{ERNIE-Doc}: A Retrospective Long-Document Modeling Transformer}

\date{}

\begin{document}
\maketitle
\begin{abstract}
\blfootnote{*indicates equal contribution.}
Transformers are not suited for processing long documents, due to their quadratically increasing memory and time consumption. Simply truncating a long document or applying the sparse attention mechanism will incur the context fragmentation problem or lead to an inferior modeling capability against comparable model sizes. In this paper, we propose \mname, a document-level language pretraining model based on Recurrence Transformers~\citep{transformer_xl}. Two well-designed techniques, namely the retrospective feed mechanism and the enhanced recurrence mechanism, enable \mname~\footnote{Source code and pre-trained checkpoints can be found at \url{https://github.com/PaddlePaddle/ERNIE/tree/repro/ernie-doc}.}, which has a much longer effective context length, to capture the contextual information of a complete document. We pretrain \mname to explicitly learn the relationships among segments with an additional document-aware segment-reordering objective. Various experiments were conducted on both English and Chinese document-level tasks. \mname 
improved the state-of-the-art language modeling result of perplexity to 16.8 on WikiText-103. Moreover, it outperformed competitive pretraining models by a large margin on most language understanding tasks, such as text classification and question answering.
\end{abstract}

\section{Introduction}

\begin{figure}[!t]
\centering
\includegraphics[width=0.8\linewidth]{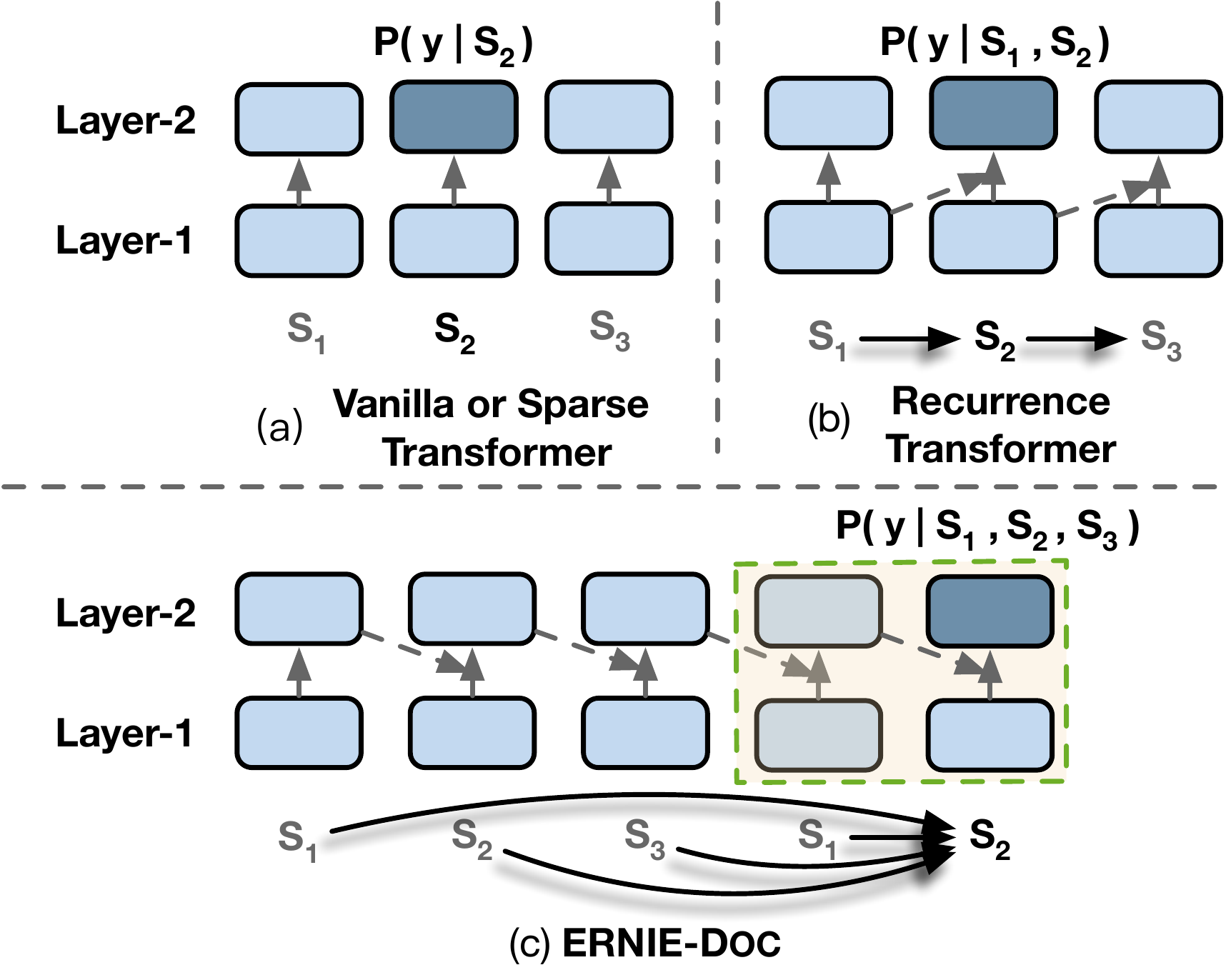}
\caption{Available contextual information utilized by Transformer variants, where a long document $\mathcal{D}$ is partitioned into three segments $\textbf{S}_i (i \in [1,2,3])$. When training on $\textbf{S}_2$, (a) and (b) optimize the pretraining objective depending only on the contextual information from the current segment or segments in the forward pass, whereas \mname utilizes the contextual information of the entire document for each segment.}
\label{fig:intro-example}
\vspace{-0.5cm}
\end{figure}

Transformers~\citep{vaswani2017attention} have achieved remarkable improvements in a wide range of natural language tasks, including language modeling~\citep{transformer_xl}, text classification~\citep{yang2019xlnet}, and question answering~\citep{devlin2018bert, radford2019language}. This success is largely due to the self-attention mechanism, which enables the network to capture contextual information from the entire input sequence. Nevertheless, the memory usage and computation complexity caused by the self-attention mechanism grows quadratically with the sequence length, incurring excessive cost when processing a long document on existing hardware.

Currently, the most prominent pretrained models, such as BERT~\citep{devlin2018bert}, are used on fixed-length input segments of a maximum of 512 tokens owing to the aforementioned limitation. Thus, a long document input must be partitioned into smaller segments of manageable sizes. However, this leads to the loss of important cross-segment information, that is, the \textit{context fragmentation} problem~\citep{transformer_xl}, as shown in Fig.~\ref{fig:intro-example}(a). To mitigate the problem of insufficient interactions among the partitioned segments of long documents, \textit{Recurrence Transformers}~\citep{transformer_xl, compressive_transformer} permit the use of contextual information from previous segments in computing the hidden states for a new segment by maintaining a memory component from the previous activation; this enables the modeling of long documents. In addition, \textit{Sparse Attention Transformers}~\citep{sparse_transformer,tay2020sparse,beltagy2020longformer,zaheer2020big} focus on reducing the complexity of self-attention operations to explicitly improve the modeling length, but only up to a restricted context length (4,096) due to resource limitations. 

We argue that existing strategies are not sufficiently effective or reliable, because \textit{\textbf{the contextual information of a complete document} is still not available for each segment during the training phase}. As depicted in Fig.~\ref{fig:intro-example}, when training on segment $S_2$, the model is ideally optimized by maximizing $P(y ~|~ (S_1, S_2, S_3))$ conditioned on the contextual information of the entire document $\mathcal{D} = \{S_1, S_2, S_3\}$, in contrast to the following suboptimal solutions: $P(y ~|~ S_2)$ for Vanilla/Sparse Transformers\footnote{For Sparse Transformers, the length of segment $S_2$ could be up to 4,096 in~\citet{beltagy2020longformer,zaheer2020big}.} and $P(y ~|~ (S_1, S_2))$ for Recurrence Transformers.

To address this limitation, we propose \mname (A Retrospective Long-Document Modeling Transformer) based on the Recurrence Transformer paradigm. Inspired by the human reading behavior of skimming a document first and then looking back upon it attentively, we design a \textbf{retrospective feed mechanism} in which segments from a document are fed twice as input. As a result, each segment in the retrospective phase could explicitly fuse the semantic information of the entire document learned in the skimming phase, which prevents context fragmentation.

However, simply incorporating the retrospective feed mechanism into Recurrence Transformers is infeasible because the maximum effective context length is limited by the number of layers~\citep{transformer_xl}, as shown in Fig.~\ref{fig:intro-example} (b). Thus, we present an \textbf{enhanced recurrence mechanism}, a drop-in replacement for a Recurrence Transformer, by changing the shifting-one-layer-downwards recurrence to the same-layer recurrence. In this manner, the maximum effective context length can be expanded, and past higher-level representations can be exploited to enrich future lower-level representations.

Moreover, we introduce a \textbf{segment-reordering objective} to pretrain a document-level model. Specifically, it is a document-aware task of predicting the correct order of the permuted set of segments of a document, to model the relationship among segments directly. This allows \mname to build full document representations for prediction. This is analogous to the sentence-reordering task in ERNIE 2.0 \citep{sun2020ernie} but at a segment level of granularity, spanning (commonly) multiple training steps. 

We first evaluate \mname on autoregressive word-level language modeling using the enhanced recurrence mechanism, which, in theory, allows the model to process a document with infinite words. \mname achieves state-of-the-art (SOTA) results on the WiKiText-103 benchmark dataset, demonstrating its effectiveness in long-document modeling. Then, to evaluate the potential of \mname on document-level natural language understanding (NLU) tasks, we pretrained the English \mname on the text corpora utilized in BigBird~\citep{zaheer2020big} from the RoBERTa-released checkpoint, and the Chinese \mname on the text corpora utilized in ERNIE 2.0 \citep{sun2020ernie} from scratch. After pretraining, we fine-tuned \mname on a wide range of English and Chinese downstream tasks, including text classification, question answering and keypharse extraction. Empirically, \mname consistently outperformed RoBERTa on various benchmarks and showed significant improvements over other high-performance long-text pretraining models for most tasks.

\section{Related Work}\label{sec:related-work}
\noindent\textbf{Sparse Attention Transformers} have been extensively explored~\citep{sparse_transformer,tay2020sparse,beltagy2020longformer,zaheer2020big}. The key idea is to sparsify the self-attention operation, which scales quadratically with the sequence length. For instance, the Sparse Transformer~\citep{sparse_transformer} uses a dilated sliding window that reduces the complexity to $ \mathcal{O}(L\sqrt{L})$, where $L$ is the sequence length. Reformer~\citep{kitaev2020reformer} further reduces the complexity to $ \mathcal{O}(L\log{L})$ using locality-sensitive hashing attention to compute the nearest neighbors. BP-Transformers ~\citep{ye2019bp} employs a binary partition for the input sequence. Recently, Longformer~\citep{beltagy2020longformer} and BigBird~\citep{zaheer2020big} have been proposed, and both achieved state-of-the-art performance on a variety of long-document tasks. They reduce the complexity of self-attention to $\mathcal{O}(L)$ by combining random attention, window attention, and global attention. However, it has been proven in ~\citet{zaheer2020big} that sparse attention mechanisms cannot universally replace dense attention mechanisms; moreover, solving the simple problem of finding the furthest vector requires $\Omega(n)$-layers of a sparse attention mechanism but only $\mathcal{O}(1)$-layers of a dense attention mechanism. In addition, the aforementioned methods require customized CUDA kernels or TVM programming to implement sparse attention, which are not maintainable and are difficult to use. In this study, we adopt a different approach to adapting Recurrence Transformers for a pretraining-then-finetuning setting, to model a long document.  

\noindent\textbf{Recurrence Transformers}~\citep{transformer_xl, compressive_transformer} have been successfully applied in generative language modeling. They employ the Transformer decoder as a parametric model for each conditional distribution in $p(\bm{x})=\prod_{t=1}^Lp(x_t|\bm{x}_{<t})$, where $\bm{x}$ denotes a text sequence. To capture long dependencies, they process the text in segments from left to right based on the segment recurrence mechanism~\citep{transformer_xl}. This mechanism maintains a memory bank of past activations at each layer to preserve a history of context. Compressive Transformer~\citep{compressive_transformer} adds a compressive memory bank to sufficiently store old activations instead of discarding them, which facilitates long-range sequence learning. However, these methods operate from left to right, which limits their capacity for discriminative language understanding tasks that require bidirectional information. XLNet~\citep{yang2019xlnet} proposed a permutation language modeling objective to construct bidirectional information and achieve superior performance in multiple NLP tasks; however, its application to long-document modeling tasks remains largely unexplored. \mname builds on the ideas of the Recurrence Transformers to 1) tackle the limitation of Recurrence Transformers for utilizing bidirectional contextual information and 2) improve the behavior of the segment recurrence mechanism to capture longer dependencies.

\noindent\textbf{Hierarchical Transformers} ~\citep{zhang2019hibert, lin2020pretrained} have enabled significant progress on numerous document-level tasks, such as document summarization \citep{zhang2019hibert} and document ranking \citep{lin2020pretrained}. Similar to Vanilla Transformers, Hierarchical Transformers also split long documents into shorter segments with manageable lengths and then feed them independently to produce corresponding segment-level semantic representations. Unlike in Vanilla Transformers, however, separate Transformer layers are used in Hierarchical Transformers to process the concatenation of these representations. Hierarchical Transformers ignore the contextual information from the remaining segments when processing each segment of a long document, thus suffering from the \textit{context fragmentation} problem.
 
\section{Proposed Method}
In this section, we first describe the background (Sec.~\ref{sec:background}) that \mname builds on. Then, we present the implementation of \mname, including the retrospective feed mechanism in Sec.~\ref{sec:recur-input}, the enhanced recurrence mechanism in Sec.~\ref{sec:lift-mem}, and the segment-reordering objective in Sec.~\ref{sec:sent-reorder}.

\subsection{Background}\label{sec:background}

\begin{figure*}[h]
\centering
\includegraphics[width=0.81\linewidth]{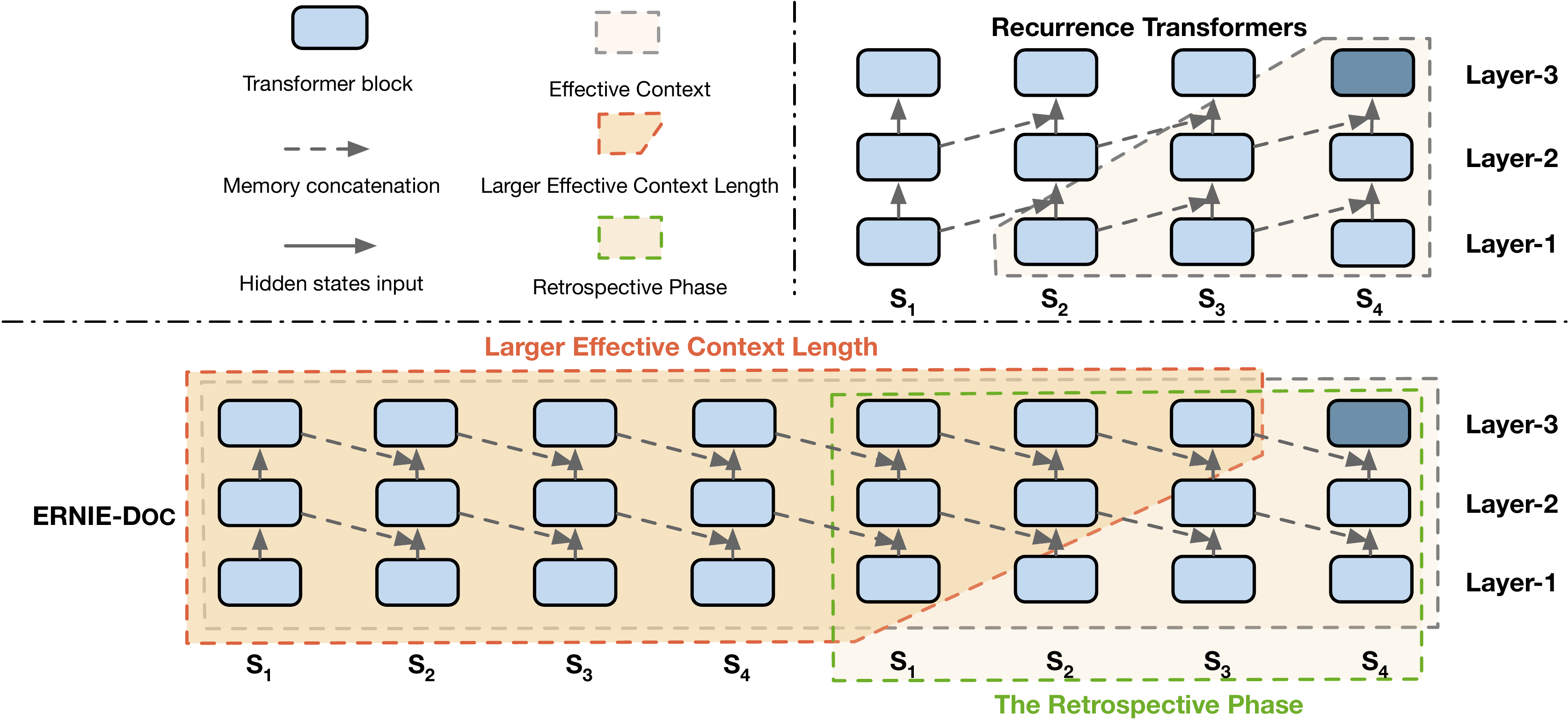}
\caption{Illustrations of \mname and Recurrence Transformers, where models with three layers take as input a long document $\mathcal{D}$ which is sliced into four segments $\textbf{S}_i, i \in [1,2,3,4]$. \textbf{Recurrence Transformers (upper-right)}: When training on $\textbf{S}_4$, it can only fuse the contextual information of the previous two consecutive segments $\textbf{S}_2,\textbf{S}_3$, since the largest effective context length grows linearly w.r.t the number of layers. \textbf{\mname (lower)}:The effective context length is much larger aided by the enhanced recurrence mechanism (Sec.~\ref{sec:lift-mem}). Thus, $\textbf{S}_4$ can fuse the information of $\textbf{S}_1$ discarded by Recurrence Transformers. Moreover, segments in the retrospective phase contains the contextual information of an entire document, powered by the retrospective feed mechanism (Sec.~\ref{sec:recur-input}).}
\label{fig:framework}
\vspace{-0.5cm}
\end{figure*}

Formally, a long document $\mathcal{D}$ is sliced into $T$ sequential segments, denoted as $\{S_1, S_2, ..., S_T\}$, where $S_{\tau}=\{x_{\tau, 1}, x_{\tau, 2}, ..., x_{\tau, L}\}$ is the $\tau$-th segment with $L$ tokens; $x$ denotes a single token. Vanilla, Sparse, and Recurrence Transformers employ different strategies to produce the hidden state $\mathbf{h}^{n}_{\tau} \in \mathbb{R}^{L \times d}$ for segment $S_{\tau}$ at the $n$-th layer:

\begin{small}
\begin{equation}
\begin{aligned}
&\widetilde{\mathbf{h}}_{\tau+1}^{n-1} = \left\{
\begin{aligned}
&\mathbf{h}_{\tau+1}^{n-1}, \text{  Vanilla or Sparse Transformers} \\
&[\mathrm{\textbf{SG}}(\mathbf{h}_{\tau}^{n-1}) \circ \mathbf{h}_{\tau+1}^{n-1}], \text{  Recurrence Transformers}, 
\end{aligned}
\right. \\
&\mathbf{q}_{\tau+1}^{n}, \mathbf{k}_{\tau+1}^{n}, \mathbf{v}_{\tau+1}^{n} =\mathbf{h}_{\tau+1}^{n-1} \mathbf{W}_{q}^{\top}, \widetilde{\mathbf{h}}_{\tau+1}^{n-1} \mathbf{W}_{k}^{\top}, \widetilde{\mathbf{h}}_{\tau+1}^{n-1} \mathbf{W}_{v}^{\top}. \\
&\mathbf{h}_{\tau+1}^{n} =\text{ Transformer-Block }(\mathbf{q}_{\tau+1}^{n}, \mathbf{k}_{\tau+1}^{n}, \mathbf{v}_{\tau+1}^{n}). 
\end{aligned}\label{eq:transformer-xl}
\end{equation}
\end{small}
where $\mathbf{q} \in \mathbb{R}^{L \times d}$, $\mathbf{k}, \text{and } \mathbf{v} \in \mathbb{R}^{(L+m) \times d}$ are the query, key and value vectors, respectively with hidden dimension $d$ and memory length $m$ (Note that $m=0$ for Vanilla or Sparse Transformers); $\widetilde{\mathbf{h}}_{\tau+1}^{n-1} \in \mathbb{R}^{(L+m) \times d}$ is the extended context; $\mathbf{W}_\ast \in \mathbb{R}^{d_\ast \times d}$ represents learnable linear projection parameters; the function \textbf{SG}$(\cdot)$ denotes the stop-gradient operation; and the notation $[\circ]$ denotes the concatenation of two hidden states along the length dimension. In contrast to Vanilla or Sparse Transformers, where $\mathbf{h}^{n}_{\tau+1}$ is produced using only itself, Recurrence Transformers introduce a segment-level recurrence mechanism to promote interaction across segments. The hidden state computed for the previous segment $\mathbf{h}^{n-1}_{\tau}$ is cached as an auxiliary context to help process the current segment $\mathbf{h}^{n}_{\tau}$. However, from the concatenation part in Eq.~\ref{eq:transformer-xl}, i.e., $[\mathrm{\textbf{SG}}(\mathbf{h}_{\tau}^{n-1}) \circ \mathbf{h}_{\tau+1}^{n-1}]$, there is apparently a constraint that the current hidden state can only fuse information from the previous segments. In other words, the contextual information of an entire document is not available for each segment.

\subsection{Retrospective Feed Mechanism}\label{sec:recur-input}

\mname employs a retrospective feed mechanism to address the unavailability of \textit{the contextual information of a complete document for each segment}. The segments from a long document are twice fed as input. Mimicking the human reading behavior, we refer to the first and second input-taking phases as the skimming and retrospective phases, respectively. In the skimming phase, we employ a recurrence mechanism to cache the hidden states for each segment. In the retrospective phase, we reuse the cached hidden states from the skimming phase to enable bi-directional information flow. Naively, we can rewrite Eq.~\ref{eq:transformer-xl} to obtain the contextual information of an entire document in the skimming phase to be utilized in the retrospective phase as follows,
\begin{equation}
\begin{aligned}
    \mathbf{\widehat{H}} &= [\mathbf{\widehat{H}}_{1:T}^{1} \circ \mathbf{\widehat{H}}_{1:T}^{2} \cdots \circ \mathbf{\widehat{H}}_{1:T}^{N}], \text{  \small (skim. phase)} \\
    \widetilde{\mathbf{h}}_{\tau+1}^{n-1} &= [ \mathrm{\textbf{SG}}(\mathbf{\widehat{H}} \circ \mathbf{h}_{\tau}^{n-1}) \circ \mathbf{h}_{\tau+1}^{n-1}], \text{  \small (retro. phase)}
\end{aligned}\label{eq:recurrent-input}
\end{equation}
where $\mathbf{\widehat{H}} \in \mathbb{R}^{(L*T*N) \times d}$ denotes the cached hidden states in the skimming phase with $T$ segments, $L$ length of each segment and total $N$ layers, and     $\mathbf{\widehat{H}}_{1:T}^i = [\mathbf{\widehat{h}}_{1}^{i} \circ \mathbf{\widehat{h}}_{2}^{i} \cdots \circ \mathbf{\widehat{h}}_{T}^{i}]$ is the concatenation of $i$-th layer's hidden states of the skimming phase. Thus, the extended context $\widetilde{\mathbf{h}}_{\tau+1}^{n-1}$ is guaranteed to capture the bidirectional contextual information of the entire document. However, it will incur massive memory and computation cost for directly employing $\mathbf{\widehat{H}}$ in self-attention mechanism. Henceforth, the main issue is how $\mathbf{\widehat{H}}$ should be implemented in a memory- and computation-efficient manner. 

By rethinking segment-level recurrence~\citep{transformer_xl}, we observe that the largest possible context dependency length increases linearly w.r.t the number of layers ($N$). For instance, at $i$-th layer, $\mathbf{\widehat{h}}_{\tau}^{i}$ have the longest dependency to $\mathbf{\widehat{h}}_{\tau-(i-1)}^{1}$. Thus, to minimize memory and computation consumption, hidden states from the $N$-th layer (top-layer) are included at a stride of $N$, which is sufficient to build the contextual information of an entire document. Formally, $\mathbf{\widehat{H}}$ can be reduced to $\mathbf{\widehat{H}}_r = [\mathbf{\widehat{h}}_{N}^{N} \circ \mathbf{\widehat{h}}_{2*N}^{N} \cdots \circ \mathbf{\widehat{h}}_{\lfloor T/N \rfloor*N}^{N}]$ (Note that when $T$ is not evenly divisible by $N$, the last hidden state  $\mathbf{\widehat{h}}_{T}^{N}$ need to be included). However, for a long document input, the extra computational and memory cost of $\mathbf{\widehat{H}}_r \in \mathbb{R}^{\lceil T / N \rceil \times d}$ where $T \gg N$ is still excessive on existing hardware.


\subsection{Enhanced Recurrence Mechanism}\label{sec:lift-mem}
To effectively utilize the retrospective feed mechanism in practice, an ideal strategy is to ensure that the cached hidden state $\mathbf{h}_\tau^{n-1}$ already contains the contextual information of an entire document without explicitly taking $\mathbf{\widehat{H}}$ or $\mathbf{\widehat{H}}_r$ as input. Essentially, we should tackle the problem of limited effective context length in the segment-level recurrence mechanisms. Herein, we introduce the enhanced recurrence mechanism, a drop-in replacement for the segment-level recurrence mechanism, by changing the shifting-one-layer-downwards recurrence to the same-layer recurrence as follows: 

\begin{equation}
   \widetilde{\mathbf{h}}_{\tau+1}^{n-1} = [\colorbox{pink}{\textbf{SG}($\mathbf{h}_{\tau}^{n}$)} \circ \mathbf{h}_{\tau+1}^{n-1}]
\label{eq:enhanced-recur}
\end{equation}
where the cached hidden state $\mathbf{h}_{\tau}^{n-1}$ in Eq.~\ref{eq:transformer-xl} and Eq.~\ref{eq:recurrent-input} is replaced with $\mathbf{h}_{\tau}^{n}$ in Eq.~\ref{eq:enhanced-recur}. 

As shown in Fig.~\ref{fig:framework}, when the retrospective feed mechanism is combined with the enhanced recurrence mechanism, every segment in the retrospective phase (shown in the box with a green dotted border) has bidirectional contextual information of the entire text input. We successfully modeled a larger effective context length (shown in the box with a orange dotted border) than traditional Recurrence Transformers can without extra memory and computation costs. Another benefit of the enhanced recurrence scheme is that past higher-level representations can be exploited to enrich future lower-level representations.

\subsection{Segment-Reordering Objective}\label{sec:sent-reorder}
\begin{figure}[t]
\centering
\includegraphics[width=\linewidth]{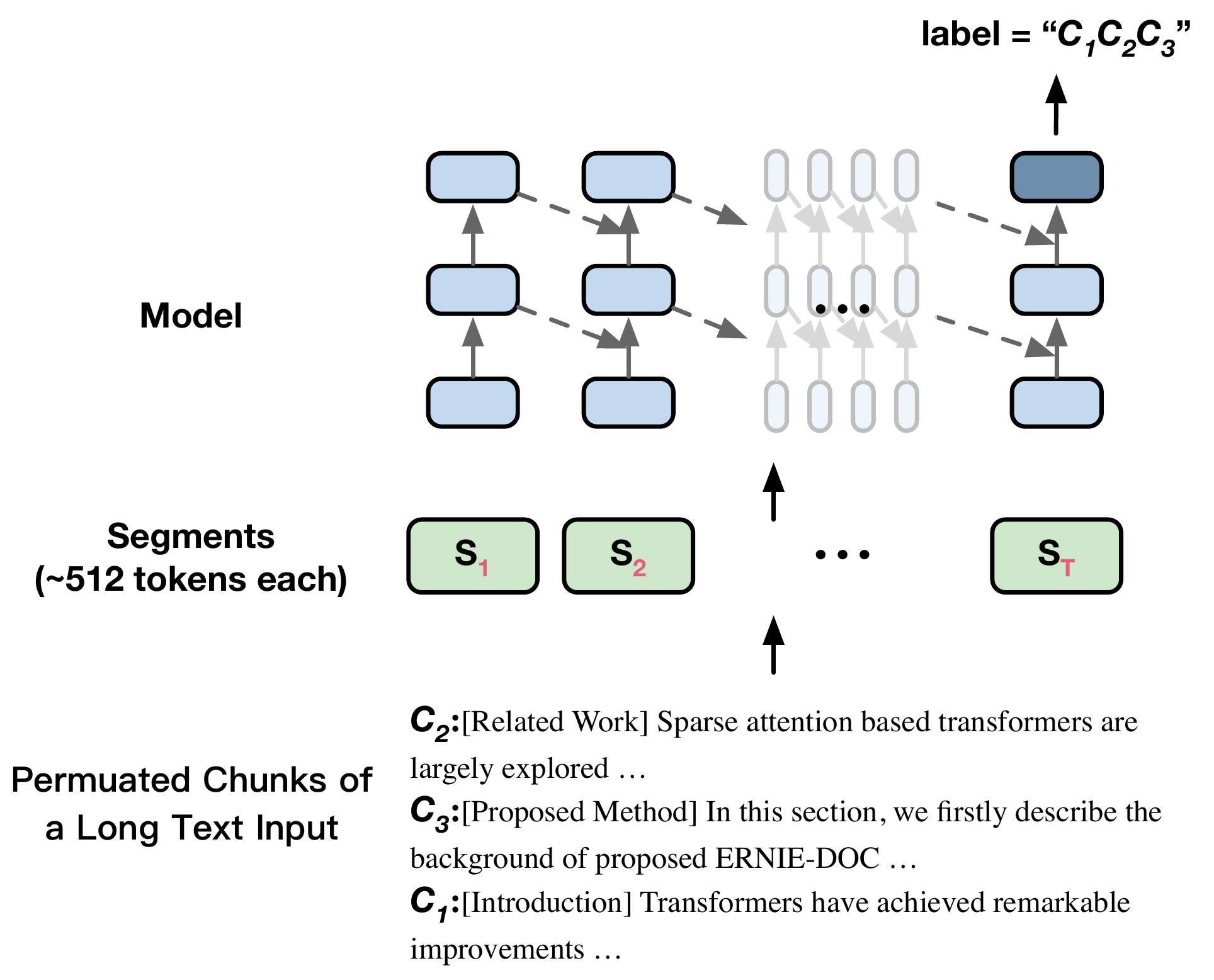}
\caption{Illustrations of segment-reordering objective.}
\label{fig:segment-reorder}
\vspace{-0.5cm}
\end{figure}

In addition to the \textbf{masked language model (MLM) objective}~\citep{devlin2018bert}, we introduce an additional document-aware task called \textbf{segment-reordering objective} for pretraining. Benefitting from the much larger effective context length provided by the enhanced recurrence mechanism, the goal of the segment-reordering objective is to predict the correct order for the permuted set of segments of a long document, to explicitly learn the relationships among segments. During the pretraining process of this task, a long text input $\mathcal{D}$ is first randomly partitioned into 1 to $m$ chunks; then, all the combinations are shuffled in a random order. As shown in Fig.~\ref{fig:segment-reorder}, $\mathcal{D}$ is partitioned into three chunks and then permuted, that is, $\mathcal{D} = \{ C_1, C_2, C_3\} \Longrightarrow \mathcal{\hat{D}} = \{ C_2, C_3, C_1\}$, where $C_i$ denotes the $i$-th chunk. Subsequently, the permuted long context $\mathcal{\hat{D}}$ is split into T sequential segments as a common practice, denoted as $\mathcal{\hat{D}}=\{S_1, S_2, ..., S_T\}$. We let the pretrained model reorganize these permuted segments, modeled as a $K$-class classification problem, where $K = \sum_{i=1}^m i!$. 

The pretraining objective is summarized as follows for the $\tau$-th input segment: 
\begin{equation}
    \max_\theta \log p_\theta(S_\tau|\hat{S_\tau}) + \mathbbm{1}_{\tau=T} \log p_\theta(\mathcal{D}|\mathcal{\hat{D}}) \nonumber
\end{equation}
where $\hat{S_\tau}$ is the corrupted version of $S_\tau$, which is obtained by randomly setting a portion of tokens to \texttt{[MASK]}; $\mathcal{\hat{D}}$ is the permutated version of $\mathcal{D}$; $\theta$ is the model parameter; and $\mathbbm{1}_{\tau=T}$ indicates that the segment-reordering objective is optimized only at the $T$-th step.

\section{Experiments}
\subsection{Autoregressive Language Modeling}
Autoregressive language modeling aims to estimate the probability distribution of an existing token/character based on previous tokens/characters in an input sequence. For comparison with previous work, we conducted experiments on word-level LM, that is, WikiText-103 \citep{wikitext}, which is a document-level language modeling dataset.

\subsubsection{Experimental Setup}
For autoregressive language modeling, we use a memory-enhanced Transformer-XL~\citep{transformer_xl}, that is, we employ our enhanced recurrence mechanism to replace the primitive one used in the Transformer-XL. Additionally, as proposed by Segatron \citep{bai2020segabert}, we introduce the segment-aware mechanism into Transformer-XL. Based on Transformer-XL, we trained a base-size model (L=16, H=410, A=10) and a large-size model (L=18, H=1,024, A=16)\footnote{We denote the number of Transformer layers as L, the hidden size as H, and the number of self-attention heads as A.}. The models were trained for 200K/400K steps using a batch size of 64/128 for the base/large configurations. During the training phase, the sequence length and memory length were limited to 150 and 384 for the base and the large model, respectively. The remaining hyper-parameters were identical to those of Transformer-XL.

\subsubsection{Results}
Tab.~\ref{tb: lm results} summarizes the evaluation results for WikiText-103. \mname achieves an impressive improvement compared with Transformer-XL: the perplexity (PPL) decreases by 3.0 for the base model and by 1.5 for the large model. Finally, we improve the state-of-the-art result of PPL to \textbf{21.0} (the base model) and \textbf{16.8} (the large model).

\begin{table}[]
\small
\setlength\tabcolsep{2pt}
\begin{tabular}{lcc}
\hline \hline
\textbf{Models}                             & \multicolumn{1}{l}{\#Param.} & \multicolumn{1}{l}{PPL} \\ \hline
{\textit{ Results of \textbf{ base models}}}          & \multicolumn{1}{l}{}         & \multicolumn{1}{l}{}    \\
LSTM \citep{grave2016improving}                               & -                            & 48.7                    \\
LSTM+Neural cache \citep{grave2016improving}                   & -                            & 40.8                    \\
GCNN-14 \citep{dauphin2017language}                           & -                            & 37.2                    \\
QRNN \citep{merity2018analysis}                              & 151M                         & 33.0                    \\
Transformer-XL Base \citep{transformer_xl}         & 151M                         & 24.0                    \\
SegaTransformer-XL Base \citep{bai2020segabert}     & 151M                         & 22.5                    \\
\mname Base & 151M                         & \textbf{21.0}           \\ \hline \hline
{\textit{ Results of \textbf{ large models}}}         &                              &                         \\
Adaptive Input \citep{baevski2018adaptive}                    & 247M                         & 18.7                    \\
Transformer-XL Large \citep{transformer_xl}       & 247M                         & 18.3                    \\
Compressive Transformer \citep{compressive_transformer}    & 247M                         & 17.1                    \\
SegaTransformer-XL Large \citep{bai2020segabert}   & 247M                         & 17.1                    \\
\mname Large & 247M                         & \textbf{16.8}                    \\ \hline
\end{tabular}
\caption{Comparison between Transformer-XL and competitive baseline results on WikiText-103.}
\label{tb: lm results}
\vspace{-0.5cm}
\end{table}

\subsection{Pretraining and Finetuning}
\subsubsection{Pretraining Text Corpora}

\begin{table}[!h]
\centering
\resizebox{0.4\textwidth}{!}{
\begin{tabular}{lrrr}
\hline \hline
Dataset & \# tokens & Avg len & Size \\ \hline
\textsc{Wikipedia}    & 2.7B      & 480         & 8G   \\
\textsc{BooksCorpus}    & 1.2B      & 2,010        & 3.5G \\
\textsc{CC-News} & 14B       & 560         & 42G  \\
\textsc{Stories} & 7.5B      & 1,891        & 22G  \\ \hline
\end{tabular}}\caption{English datasets used for pretraining.}
\label{tb:english_pretraining_data}
\end{table}

\noindent{\textbf{English Data.}} To allow \mname to capture long dependencies in pretraining, we compiled a corpus from four standard datasets: \textsc{Wikipedia}, \textsc{BooksCorpus}~\citep{zhu2015aligning}, \textsc{CC-News}\footnote{We used \texttt{news-please} to crawl English news articles published between September 2016 and February 2019 and adopted 
Message Digest Algorithm5 (MD5) for deduplication.}, and \textsc{Stories}~\citep{trinh2018simple} (details listed in Tab.~\ref{tb:english_pretraining_data}). We tokenized the corpus using the RoBERTa wordpieces tokenizer \citep{liu2019roberta} and duplicated the pretraining data 10 times.  
\\ 
\noindent{\textbf{Chinese Data.}} The Chinese text corpora used in ERNIE 2.0 \citep{sun2020ernie} were adopted for pretraining \mname.

\subsubsection{Experimental Setup} 

\noindent{\textbf{Pretraining.}} We trained three sizes of models for English tasks: small (L=6, H=256, A=4), base (L=12, H=768, A=12), and large (L=24, H=1,024, A=16). For Chinese tasks, we used only one size, i.e., base (L=12, H=768, A=12). We limited the length of the sentences in each mini-batch to 512 tokens and the length of the memory to 128. The models were trained for 500K/400K/100K steps using a batch size of 2,560/2,560/3,920 sentences for the small/base/large configurations. \mname was optimized with the Adam \citep{kingma2014adam} optimizer. The learning rate was warmed up over the first 4,000 steps to a peak value of 1e-4, and then it linearly decayed. The remaining pretraining hyperparameters were the same as those of RoBERTa \citep{liu2019roberta} (see Tab.~\ref{tab: pretraing_params}). Additionally, we employed relative positional embedding \citep{shaw2018self} in our model pretraining because it is necessary for reusing hidden state without causing temporal confusion \citep{transformer_xl}.
\\

\noindent{\textbf{Finetune.}} In contrast to previous models, such as BERT, RoBERTa, and XLNet, the proposed model employs the retrospective feed mechanism and the enhanced recurrence mechanism during the fine-tuning phase to fully utilize the advantages of these two strategies.

\subsubsection{Results on English Tasks}
\begin{table}[]
\centering
\resizebox{0.47\textwidth}{!}{
\begin{tabular}{lccc}
\hline \hline
\textbf{Models}          & \multicolumn{2}{c}{\textbf{IMDB}} & \textbf{HYP} \\ \hline
                & Acc.         & F1         & F1            \\ \hline
RoBERTa~\citep{liu2019roberta}    & 95.3        & 95.0       & 87.8          \\
Longformer~\citep{beltagy2020longformer} & 95.7        & -          & 94.8          \\
BigBird~\citep{zaheer2020big}    & -           & 95.2       & 92.2          \\
\mname       & \textbf{96.1}        & \textbf{96.1}       & \textbf{96.3}         \\ \hline
XLNet-Large~\citep{yang2019xlnet}     & 96.8        & -          & -             \\
\mname-Large      & \textbf{97.1}        & \textbf{97.1}      & \textbf{96.6}          \\ \hline
\end{tabular}}\caption{Results on the IMDB and HYP dataset for long-text classification.}
\label{tab:long-text-class-result}
\vspace{-0.5cm}
\end{table}

\noindent \textbf{Results on Long-Text Classification Tasks}. We consider two datasets: IMDB reviews ~\citep{maas2011learning} and Hyperpartisan News Detection (HYP)~\citep{kiesel2019semeval}. The former is a widely used sentiment analysis dataset containing 50,000 movie reviews, labeled as positive or negative. The latter contains news that takes extreme left-wing or right-wing standpoints. The documents in HYP are extremely long (50\% of the samples contain more than 537 tokens) and are thus suitable for testing long-text classification ability. Tab.~\ref{tab:long-text-class-result} summarizes the results of the \mname-Base and \mname-Large models for long-text classification tasks, and \mname achieves a SOTA result. On IMDB, we observed a modest performance gain compared with RoBERTa. This is because nearly 90\% of the samples in the dataset consist of fewer than 569 tokens. Unlike on IMDB, \mname surpasses the baseline models on HYP by a substantial margin, demonstrating its capability of utilizing information from a long document input. Note that we include XLNet-Large, the previous SOTA pretraining model on the IMDB dataset, as the baseline for a large model setting; \mname achieves a result comparable to that of XLNet-Large. 

\noindent \textbf{Results on Document-level Question-Answering Tasks}.
\begin{table}[]
\centering
\resizebox{0.43\textwidth}{!}{
\begin{tabular}{lcccc}
\hline \hline
\textbf{Models}  & \textbf{TQA} & \multicolumn{3}{c}{\textbf{HQA}}  \\ \hline
                & F1       & Span & Supp & Joint \\ \hline
RoBERTa        & 74.3     & 73.5    & 83.4     & 63.5     \\
Longformer     & 75.2     & 74.3    & 84.4     & 64.4     \\
BigBird        & 79.5     & 75.5    & \textbf{87.1}     & 67.8     \\
\mname      & \textbf{80.1}        & \textbf{79.4}    & 86.3     & \textbf{70.5}     \\
\hline
Longformer-Large    & 77.8        & 81.0          & 85.8 & 71.4            \\
BigBird-Large & - & 81.3 & \textbf{89.4} & - \\
\mname-Large      & \textbf{82.5}        & \textbf{82.2}      & 87.6 & \textbf{73.7}        \\ \hline
\end{tabular}}\caption{Results on TQA and HQA dev dataset for document-level QA. HQA metrics are F1.}\label{tab:qa-result}
\end{table}
We utilized two document-level QA datasets (Wikipedia setting of TriviaQA (TQA)~\citep{joshi2017triviaqa} and distractor setting of HotpotQA (HQA)~\citep{yang2018hotpotqa}) to evaluate the reasoning ability of the models over long documents. TQA and HQA are extractive QA tasks, and we follow the simple QA model of BERT~\citep{devlin2018bert} to predict an answer with the maximum sum of start and end logits across multiple segments of a sample. In addition, we use a modified cross-entropy loss~\citep{clark2017simple} for the TQA dataset and use a two-stage model~\citep{groeneveld2020simple} with the backbone of \mname for the HQA dataset. Tab.~\ref{tab:qa-result}. 
shows that \mname outperforms RoBERTa and Longformer by a considerable margin on these two datasets, and is comparable to current SOTA long-document model, i.e., BigBird on HQA in large-size model setting.


\begin{table}[]
\centering
\resizebox{0.47\textwidth}{!}{
\begin{tabular}{lccc}
\hline \hline
\textbf{OpenKP dataset}                 & F1@1 & F1@3 & F1@5 \\ \hline
BLING-KPE~\citep{xiong2019open} & 26.7 & 29.2 & 20.9 \\  
JointKPE~\citep{sun2020joint} & 39.1     & 39.8     & 33.8     \\
ETC~\citep{ainslie2020etc}              & -     & 40.2     & -     \\
\mname             & \textbf{40.2}     & \textbf{40.5}     & \textbf{34.4}     \\ \hline
\end{tabular}}\caption{Results on OpenKP dev dataset. The baseline results are obtained from corresponding papers under no-visual-features setting.}\label{tab:kp-result}
\vspace{-0.5cm}
\end{table}

\noindent \textbf{Results on the Keyphrase Extraction Task}. We include OpenKP~\citep{xiong2019open} dataset to evaluate \mname's ability to extract keyphrases from a long document. Each document contains up to three short keyphrases and we follow the model setting of JointKPE~\citep{sun2020joint} and ETC~\citep{ainslie2020etc} by applying CNNs on BERT's output to compose n-gram embeddings for classification. We report the results of base-size models in Tab.~\ref{tab:kp-result} under no-visual-features setting for easy and fair comparison with baselines. \mname performs stably better on all metrics on the OpenKP dataset.

\begin{table*}[]
\setlength\tabcolsep{3pt}
\resizebox{1\textwidth}{!}{
\begin{tabular}{l|ccccccccccc}
\hline \hline
                & \multicolumn{2}{c}{\textbf{DRCD}}                 & \textbf{CMRC2018}  & \textbf{DuReader}  & \multicolumn{2}{c}{\textbf{CAIL}}        & \multicolumn{2}{c}{\textbf{THU}}    & \textbf{IFK} & \multicolumn{2}{c}{\textbf{C$^3$}} \\
\textbf{Models}  & \multicolumn{2}{c}{EM/F1}                         & EM/F1              & EM/F1              & \multicolumn{2}{c}{Acc.}                  & \multicolumn{2}{c}{Acc.}                  & Acc.              & \multicolumn{2}{c}{Acc.}         \\ \cline{2-12} 
                & Dev                           & Test                              & Dev                & Dev           &Dev  & \multicolumn{1}{l}{Test} & Dev           & \multicolumn{1}{l}{Test} & Dev              & Dev            & Test           \\ \hline
BERT \citep{devlin2018bert}           & 85.7/91.6                     & 84.9/90.9          & 66.3/85.9          & 59.5/73.1          & 61.9          & 67.3                     & 97.7          & 97.3                     & 60.3             & 65.7           & 64.5           \\
BERT-wwm-ext$^*$    & 85.0/91.2                     & 83.6/90.4          & 67.1/85.7          & -/-                & -             & -                        & 97.6          & 97.6                     & 59.4             & 67.8           & 68.5           \\
RoBERTa-wwm-ext$^*$  & \multicolumn{1}{l}{86.6/92.5} & 85.2/92.0          & 67.4/87.2          & -/-                & -             & -                        & -             & -                        & 60.3             & 67.1              & 66.5              \\
MacBERT \citep{cui2020revisiting}        & 88.3/93.5                     & 87.9/93.2          & 69.5/87.7          & -/-                & -             & -                        & -             & -                        & -                & -              & -              \\
XLNet-zh \citep{cui2020}        & 83.2/92.0                     & 82.8/91.8          & 63.0/85.9          & -/-                & -             & -                        & -             & -                        & -                & -              & -              \\
ERNIE 1.0 \citep{sun2019ernie}      & 84.6/90.9                     & 84.0/90.5          & 65.1/85.1          & 57.9/72/1          & -             & -                        & 97.7          & 97.3                     & 59.0             & 65.5           & 64.1           \\
ERNIE 2.0 \citep{sun2020ernie}      & 88.5/93.8                     & 88.0/93.4          & 69.1/88.6          & 61.3/74.9          & 64.9          & 67.9                     & 98.0          & 97.5                     & 61.7             & 72.3           & 73.2           \\ \hline
\mname      & \textbf{90.5/95.2}            & \textbf{90.5/95.1} & \textbf{76.1/91.6} & \textbf{65.8/77.9} & \textbf{65.6} & \textbf{68.8}            & \textbf{98.3} & \textbf{97.7}            & \textbf{62.4}    & \textbf{76.5}  & \textbf{76.5}  \\ \hline
\end{tabular}
}
\caption{Results on seven Chinese NLU tasks for \mname-base model.The results of the models with "$\ast$" are from \citet{cui2019pre}. The XLNet-zh is the abbreviation of Chinese-XLNet. Notably, the result of BERT on CAIL was obtained from \citet{cail}, where BERT was post-pretrained with a legal dataset.}
\label{tb: chinese results}
\end{table*}

\subsubsection{Results on Chinese Tasks}
We conducted extensive experiments on seven Chinese natural language understanding (NLU) tasks, including machine reading comprehension (CMRC2018~\citep{cmrc2018}, DRCD~\citep{drcd}, DuReader~\citep{dureader}, C$^3$~\citep{c3}), semantic similarity (CAIL2019-SCM (CAIL)~\citep{cail}), and long-text classification (IFLYTEK (IFK)~\citep{iflytek}, THUCNews (THU)\footnote{We use a subset of THUCNews (\url{https://github.com/gaussic/text-classification-cnn-rnn}).}~\citep{thucnews}). The documents in all the aforementioned datasets are sufficiently long to be used to evaluate the effectiveness of \mname on long-context tasks (see detailed datasets statistics in Tab.~\ref{tab:ch-dataset-stat}). We reported the mean results with five runs for the seven Chinese tasks in Tab.~\ref{tb: chinese results}, and summarized the hyperparameters in Tab.~\ref{tab:params_nlu_zh}. \mname outperforms previous models across these Chinese NLU tasks by a significant margin in the base-size model group.  

\subsubsection{Ablation Studies}
\begin{table}[H]
\centering
\resizebox{0.48\textwidth}{!}{
\begin{tabular}{llcl}
\hline \hline
\textbf{No.} & \textbf{Models}                & \textbf{TQA}   & \textbf{HYP}   \\ \hline
\textbf{\uppercase\expandafter{\romannumeral1}}   & \mname & 64.56 & 86.10 \\
\textbf{\uppercase\expandafter{\romannumeral2}}   & \uppercase\expandafter{\romannumeral1} w/o segment-reordering              & 63.59 & 84.60 \\
\textbf{\uppercase\expandafter{\romannumeral3}}   & \uppercase\expandafter{\romannumeral2} w/o retrospective feed           & 63.38 & 83.27 \\
\textbf{\uppercase\expandafter{\romannumeral4}}   & \uppercase\expandafter{\romannumeral3} w/o enhanced recurrence          & 61.09 & 81.67 \\
\textbf{\uppercase\expandafter{\romannumeral5}}   & \uppercase\expandafter{\romannumeral4} w/o recurrence           & 58.35 & 77.72 \\ \hline
\end{tabular}}\caption{Performance of \mname-Small after ablating each proposed component (F1 result is reported).}\label{tab:ab-small-result}
\vspace{-0.5cm}
\end{table}

\noindent\textbf{Effect of proposed components}. Tab.~\ref{tab:ab-small-result} shows the performance of \mname-Small on two English tasks after ablating each proposed component. All models were pretrained and fine-tuned with the same experimental setup, and we report the mean results of five runs. We observed a stable performance gain across these two tasks by incorporating each proposed component. By comparing No.\uppercase\expandafter{\romannumeral4} and No.\uppercase\expandafter{\romannumeral5}, we see that segment-level recurrence is necessary for modeling long documents and produces 2.74 and 3.95 \% points improvement on the TQA and HYP dateset, respectively. Moreover, a substantial improvement is achieved using the enhance recurrence mechanism (2.29\% point on TQA and 1.40\% point on HYP, see No.\uppercase\expandafter{\romannumeral3} - \uppercase\expandafter{\romannumeral4}). Retrospective feed mechanism further improves 0.21\% point on TQA and 1.33\% point on HYP (No.\uppercase\expandafter{\romannumeral2} - No.\uppercase\expandafter{\romannumeral3}). Considering different types of tasks, we observe that on HYP, an extremely long text classification dataset, a substantial improvement is achieved using the segment-reordering objective (1.5\% point). This indicates that the \texttt{[CLS]} token, pretrained using the segment-reordering objective, is more adaptable to the document-level text classification task.
\begin{figure}[H]
\centering
\includegraphics[width=\linewidth]{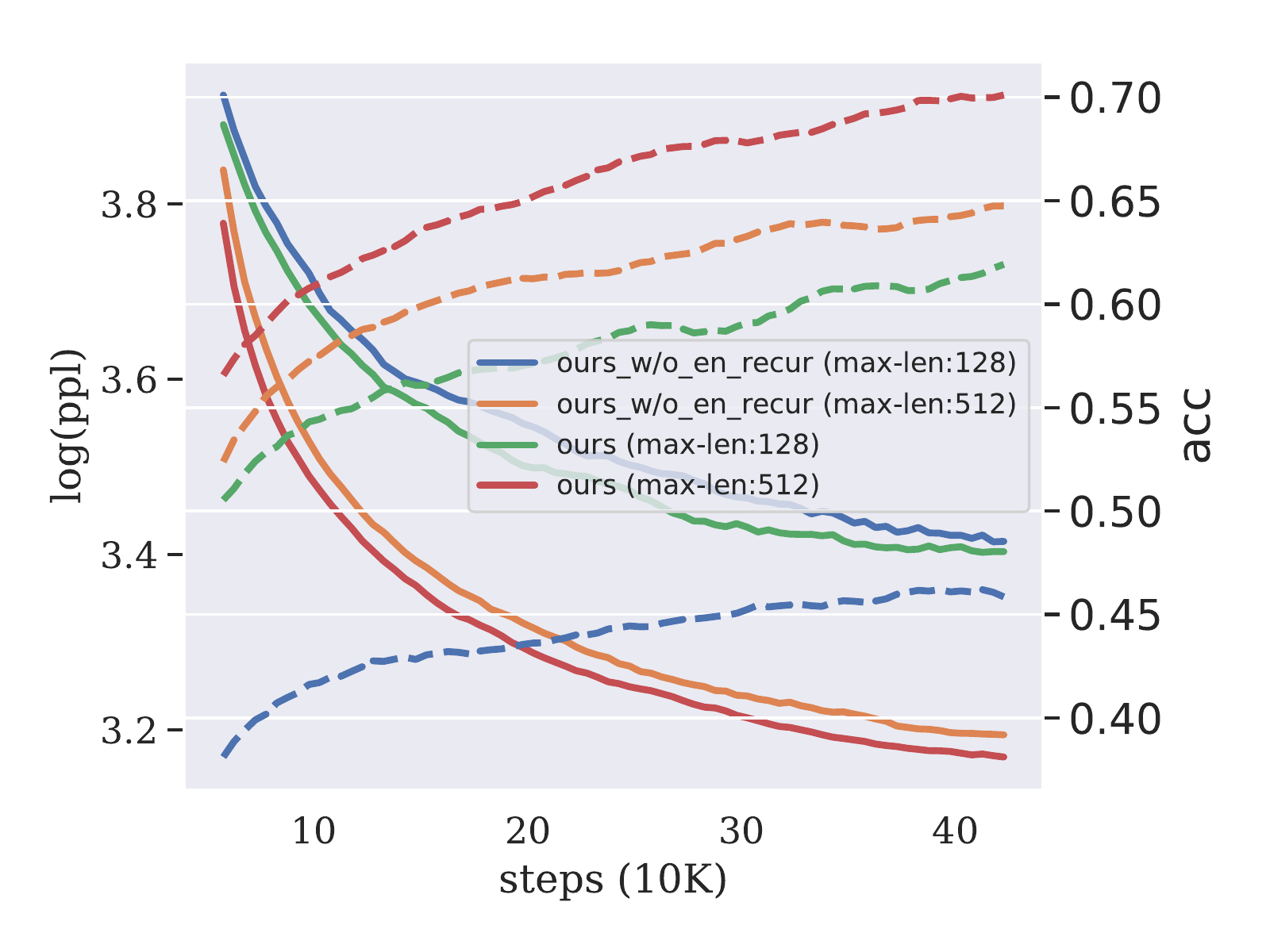}
\vspace{-1cm}
\caption{Acc. (dotted line) and PPL (solid line) metrics for variants of our small models with different maximum sequence length during pretraining.}
\label{fig:small_acc_ppl}
\vspace{-0.5cm}
\end{figure}

\noindent\textbf{Effect of enhanced recurrence mechanism with regard to different maximum sequence lengths}. As depicted in Fig.~\ref{fig:small_acc_ppl}, the enhanced recurrence mechanism plays an important role in pretraining an effective language model with lower PPL and higher accuracy under both the maximum sequence input lengths of 128 and 512. 
The effect of the enhanced recurrence mechanism is more significant under a smaller maximum sequence length, even makes the \mname-Small (max-len:128) comparable to \mname-Small$\_$w/o$\_$en$\_$recur (max-len:512) w.r.t accuracy. This intriguing property of the enhanced recurrence mechanism enables more efficient model training and inference by reducing maximum sequence length while remaining comparable modeling capability.


\section{Conclusion}
In this paper, we proposed \mname, a document-level language pretraining model based on the Recurrence Transformers paradigm. Two well-designed mechanisms, namely the retrospective feed mechanism and the enhanced recurrent mechanism, enable \mname, which theoretically has the longest possible dependency, to model bidirectional contextual information of a complete document. Additionally, \mname is pretrained with a document-aware segment-reordering objective to explicitly learn the relationship among segments of a long context. Experiments on various downstream tasks demonstrate that \mname outperforms existing strong pretraining models such as RoBERTa, Longformer, and BigBird and achieves 
SOTA results on several language modeling and language understanding benchmarks.\\
In future studies, we will evaluate \mname on language generation tasks, such as generative question answering and text summarization. We will also investigate its potential applicability in other areas, such as computational biology. Another possibility is to incorporate graph neural networks into \mname to enhance its modeling capability for tasks that require multi-hop reasoning and long-document modeling ability.

\section*{Acknowledgments}
This work was supported by the National Key Research and Development Project of China (No. 2018AAA0101900).

\bibliography{anthology,acl2021}
\bibliographystyle{acl_natbib}

\newpage

\appendix
\section{Appendices}
\label{sec:appendix}
\subsection{Tasks}
\begin{table*}[h]
\centering
\small
\setlength\tabcolsep{7.5pt}
\resizebox{1\textwidth}{!}{
\begin{tabular}{lccccccccccc}
\hline \hline
\textbf{Datasets}                         & \multicolumn{2}{c}{\textbf{IMDB}} & \multicolumn{3}{c}{\textbf{Hyperpartisan}} & \multicolumn{2}{c}{\textbf{TriviaQA}} & \multicolumn{2}{c}{\textbf{HotpotQA}} & \multicolumn{2}{c}{\textbf{OpenKP}}\\ \hline
\multicolumn{1}{l|}{split}      & train       & dev        & train      & dev      & test              & train          & dev         & train         & dev & train & dev         \\
\multicolumn{1}{l|}{\# samples} & 25,000       & 2,000      & 516        & 64       & 65                &  61,888              & 7,993            & 90,432         & 7,404 & 134,894 & 6,616       \\ \hline
\multicolumn{12}{l}{\# tokens of context length in each percentile using RoBERTa wordpiece tokenizer}                                                                                      \\ \hline
\multicolumn{1}{l|}{50\%}       & 215         & 212        & 537        & 521      & 639             &  8,685              & 8,586            & 1,279          & 1,325 & 894 & 681       \\
\multicolumn{1}{l|}{90\%}       & 569         & 550        & 1,519       & 1,539     & 1,772              &  25,207              & 24,825            & 1,725          & 1,785  & 3,451 & 2,734      \\
\multicolumn{1}{l|}{95\%}       & 745         & 724        & 1,997       & 1,979     & 1,994              &  32,018              & 32,132            & 1,888          & 1,943  & 5,340 &  4,130     \\
\multicolumn{1}{l|}{max}        & 3,084        & 2,778       & 5,566       & 2,643     & 5,566             &  173,302              & 146,012            & 3,733          & 3,618  & 105,548 & 43,609      \\ \hline
\end{tabular}
}
\caption{English Datasets statistics.}
\label{tab:nlu-dataset-stat}
\end{table*}

\begin{table*}[h]
\centering
\small
\setlength\tabcolsep{3.5pt}
\resizebox{1\textwidth}{!}{
\begin{tabular}{lccccccccccccccc}
\hline \hline
\textbf{Datasets}               & \multicolumn{2}{c}{\textbf{IFLYTEK}} & \multicolumn{2}{c}{\textbf{THUCNews}} & \multicolumn{2}{c}{\textbf{CAIL}} & \multicolumn{2}{c}{\textbf{CMRC2018}} & \multicolumn{2}{c}{\textbf{DuReader}} & \multicolumn{2}{c}{\textbf{C$^3$}} & \multicolumn{3}{c}{\textbf{DRCD}} \\ \hline
\multicolumn{1}{l|}{split}      & train             & dev              & train              & dev              & train            & dev            & train              & dev              & train              & dev              & train           & dev           & train      & dev       & test     \\
\multicolumn{1}{l|}{\# samples} & 12,133             & 2,599             & 50,000              & 5,000             & 5,102             & 1,500           & 10,121              & 3,219             & 15,763              & 1,628             & 11,869           & 3,816          & 26,936      & 3,524      & 3,493     \\ \hline
\multicolumn{16}{l}{\# tokens of context length in each percentile using BERT tokenizer}                                                                                                                                                                                                        \\ \hline
\multicolumn{1}{l|}{50\%}       & 243               & 242              & 656               & 579             & 1,837             & 1,834           & 423                & 426              & 163                & 182              & 96             & 89             & 397        & 421       & 405      \\
\multicolumn{1}{l|}{90\%}       & 507             & 508              & 1,821              & 1,599              & 1,965             & 1,962           & 745                & 771              & 550                & 567              & 591          & 554            & 616        & 666       & 626      \\
\multicolumn{1}{l|}{95\%}       & 563             & 560            & 2,455              & 2,245              & 2,008             & 1,995           & 827                & 840              & 652                & 667              & 697            & 692          & 709        & 740       & 736      \\
\multicolumn{1}{l|}{max}        & 3,153              & 1,698             & 26,659             & 9,128              & 2,400             & 2,310           & 970                & 961              & 1,021               & 854              & 1,534           & 1,167           & 1,678       & 989       & 950      \\ \hline

\end{tabular}
}
\caption{Chinese Datasets statistics.}
\label{tab:ch-dataset-stat}
\vspace{-0.5cm}
\end{table*}

Following previous work, we evaluate \mname on various tasks that require the ability to model a long document.
\\

\noindent \textbf{Document-level Language Modeling Task.} We employ WikiText-103 \citep{wikitext} in language modeling experiments. WikiText-103 is the largest available word-level benchmark with long-term dependency for language modeling, which consists of 28K articles, where each article has 3.6K tokens on average, thus 103M training tokens in total.
\\

\noindent \textbf{Long Text classification}. We consider two English datasets: IMDB reviews ~\citep{maas2011learning} and Hyperpartisan news detection~\citep{kiesel2019semeval} (see Tab.~\ref{tab:nlu-dataset-stat}), and two Chinese datasets: IFLYTEK \citep{iflytek} and THUCNews \citep{thucnews} (see Tab.~\ref{tab:ch-dataset-stat}). 
IMDB is a widely used sentiment analysis dataset containing 50,000 movie reviews labeled as positive or negative. Training and dev dataset is equally split. 
Hyperpartisan contains news that takes an extreme left-wing or right-wing standpoint. Documents are extremely long in Hyperpartisan which makes it a good test for long text classification. We use the same split as Longformer by dividing 654 documents into train/dev/test sets.
IFLYTEK contains 17,332 app descriptions. The task is to assign each description into one of 119 categories, such as food, car rental and education.
THUCNews is generated by filtering historical data of Sina News RSS subscription channel from 2005 to 2011, including 740,000 news documents and 14 categories. In this paper, we employ the subset version instead of the full one \footnote{The subset version is also released and can be downloaded from the official website of THUCTC.}, which contains 10 categories, each with 5,000 pieces of data.


For the above four long text classification datasets, we concatenate [CLS] token with each segment and takes as input multiple segments of a text sequentially. Each segment is generated by slicing the text with a sliding window of 128 tokens. We apply binary cross entropy loss on the [CLS] token of the last segment.
\\

\noindent \textbf{Long Text Semantic Similarity.} Considering that there is no available long text semantic similarity dataset in English, we evaluate the effectiveness of \mname on semantic similarity task only depending on Chinese dataset CAIL2019-SCM. According to \citet{cail}, CAIL2019-SCM is a sub-task of the \textbf{C}hinese \textbf{AI} and \textbf{L}aw Challenge (CAIL) competition in 2019, which contains 8,964 triplets of legal documents collected from China Judgments Online. Every document in a majority of triplet has more than 512 characters, therefore, the total length of a triplet is quite long. CAIL2019-SCM requires researchers to decide which two cases are more similar in a triplet. Specifically, given a triplet $(A, B, C)$, where A, B, C are fact descriptions of three cases. The model needs to predict whether $sim(A, B) > sim(A, C)$ or $sim(A, C) > sim(A, B)$, in which $sim$ denotes the similarity between two cases. Instead of separately feeding the document A, B, C into the model to get the feature $h$, we use the combinations of $(A, B)$ and $(A, C)$ as input. We generate multiple segments for $(A, B)$ or $(A, C)$ with a sliding window of 128 tokens and feed them as input sequentially. The binary cross entropy loss is applied to the difference of [CLS] token output of each segment.
\\

\begin{table*}[h]
\centering
\resizebox{0.7\textwidth}{!}{
\begin{tabular}{lccccc}
\hline \hline
                              & \multicolumn{2}{c}{QA} & \multicolumn{2}{c}{Classification} &       \\ \cline{2-6} 
\textbf{Models}\textbackslash{}\textbf{Dataset} & \textbf{TriviaQA}   & \textbf{HotpotQA}  & \textbf{IMDB}  & \textbf{Hyperpartisan}              & \textbf{Avg.}  \\ \hline
\#0 \mname     & 64.56      & 50.85     & 93.14 & \multicolumn{1}{c|}{86.10} & \textbf{73.66} \\
\#1 w/o so                    & 63.59      & 50.04     & 93.15 & \multicolumn{1}{c|}{84.60} & 72.85 \\
\#2 w/o so\&retro             & 63.38      & 49.87     & 92.56 & \multicolumn{1}{c|}{83.27} & 72.27 \\
\#3 w/o so\&retro\&en-rec     & 61.09      & 44.05     & 92.07 & \multicolumn{1}{c|}{81.67} & 69.72 \\
\#4 w/o so\&retro\&recur      & 58.35      & 31.54     & 91.60 & \multicolumn{1}{c|}{77.72} & 64.80 \\ \hline
\end{tabular}}\caption{Performance of \mname-small after ablating each proposed component. (\textbf{so} denotes the segment-reordering objective, \textbf{re} denotes the retrospective feed mechanism, \textbf{en-rec} denotes the enhanced recurrence mechanism, and \textbf{recur} denotes the segment-level recurrence module. We used the Acc. metric for IMDB, F1 metric for TriviaQA and Hyperpartisan, Joint-F1 for HotpotQA.)}\label{tab:ab-finetuing}
\vspace{-0.5cm}
\end{table*}

\noindent \textbf{Document-level Question answering}. We utilize two English question answering datasets (TriviaQA~\citep{joshi2017triviaqa}, HotpotQA~\citep{yang2018hotpotqa}) (see Tab.~\ref{tab:nlu-dataset-stat}) and four Chinese question answering datasets (CMRC2018~\citep{cmrc2018}, DRCD~\citep{drcd}, DuReader~\citep{dureader}, C$^3$~\citep{c3}) (see Tab.~\ref{tab:ch-dataset-stat}) to evaluate models' reasoning ability over long documents. 

TriviaQA is a large scale QA dataset that contains over 650K question-answer pairs. We evaluate models on its Wikipedia setting where documents are Wikipedia articles, and answers are named entities mentioned in multiple documents. The dataset is distantly supervised meaning that there is no golden span, thus we find all superficial identical answers in provided documents\footnote{We use the same preprocessing code for TriviaQA dataset as BigBird, see \url{https://github.com/tensorflow/models/blob/master/official/nlp/projects/triviaqa/preprocess.py}}. We use the following input format for each segment: ``\texttt{[CLS] context [q] question [/q]}'' where context is generated by slicing multi-documents input with a sliding window of 128 tokens. We take as input multiple segments of a sample sequentially and attach a linear layer to each token in a segment to predict the answer span. We use a modified cross entropy loss \citep{clark2017simple} assuming that each segment contains at least one correct answer span. The final prediction for each question is a span with the maximum sum of start and end logit across multiple segments. 

HotpotQA is a QA dataset where golden spans of an answer and sentence-level supporting facts are provided. Thus, it contains two tasks namely, answer span prediction and supporting facts prediction. In the distractor setting, each question is associated with 10 documents where only 2 documents contain supporting facts. It requires the model to find and reason over multiple documents to find answers, and explain the predicted answers using predicted supporting facts. Following~\citet{groeneveld2020simple}, we implemented a two-stage model based on \mname and use the following input format for each segment: ``\texttt{[CLS] title$_1$ [p] sent$_{1,1}$ [SEP] sent$_{1,2}$ [SEP] ... title$_2$ [p] sent$_{2,1}$ [SEP] sent$_{2,2}$ [SEP] ... [q] question [/q]}'' For evidence prediction, we apply 2 layer feedforward networks over the special token [SEP] and [p] representing a sentence and a paragraph separately. Then we use binary cross entropy loss to do binary classification. For answer span prediction, we train the model with a multi-task objective: 1) question type (yes/no/span) classification on the [CLS] token. 2) supporting evidence prediction on [SEP] and [p]. 3) span prediction on the start and end token of a golden span.

CMRC2018, DRCD and DuReader are common Chinese QA datasets with same format, which have been evaluated in numerous popular pretraining models, such as BERT \citep{devlin2018bert}, ERNIE 1.0 \citep{sun2019ernie}, ERNIE 2.0 \citep{sun2020ernie} and etc. The detailed descriptions of three datasets can refer to \citet{cmrc2018}, \citet{drcd} and \citet{dureader}. We adopt the same input format as TriviaQA for each segment, denotes as ``\texttt{[CLS] context [SEP] question [SEP]}`` where context is generated by slicing multi-documents input with a sliding window of 128 tokens. We take as input multiple segments of a sample sequentially and attach a linear layer to each token in a segment to predict the answer span. Then, we apply a softmax and use the cross entropy loss with the correct answer. The final prediction for each question is a span with the maximum sum of start and end logit across multiple segments. 

The multiple \textbf{C}hoice \textbf{C}hinese machine reading \textbf{C}omprehension dataset (C$^3$) \citep{c3} is the first Chinese free-form multi-choice dataset where each question is associated with at most four choices and a single document. According to \citep{c3}, $m$ segments are constructed for a question, in which $m$ denotes the number of choice for that question. We use the following input format for each segment:  ``\texttt{[CLS] context [SEP] question [SEP] choice$_i$ [SEP]} '' where context is generated by slicing document input with a sliding window of 128 tokens stride. We take as input multiple segments of a sample in a single batch and attach a linear layer to \texttt{[CLS]} that outputs an unnormalized logit. Then we obtain the final prediction for a question by applying a softmax layer over the unnormalized logits of all choices associated with it. 
\\

\noindent \textbf{Keyphrase Extraction}. We include OpenKP~\citep{xiong2019open} dataset~\footnote{The dataset can be downloaded from \url{https://github.com/thunlp/BERT-KPE}} to evaluate \mname's ability to extract keyphrases from a long document. Each document contains up to three short keyphrases and we follow the model setting of JointKPE~\citep{sun2020joint} and ETC~\citep{ainslie2020etc} by applying CNNs on BERT's output to compose n-gram embeddings for classification. We clean the dataset by removing some nonsense words such as the HTTP links. In detail, we apply five CNNs on BERT's output with the kernel size ranging from 1 to 5. Since each word is composed of several sub-tokens, we take the first token's embedding as the input for CNNs. Finally, we use the binary cross entropy loss as the optimization objective.

\subsection{Ablation Studies}

Tab.~\ref{tab:ab-finetuing} shows the performance of \mname-Small on English tasks after ablating each proposed component. All models were pretrained and fine-tuned with the same experimental setup, and we report the mean results of five runs. In the last column in Tab.~\ref{tab:ab-finetuing}, we see that the segment-reordering objective is improved \mname by 0.81\% on average (\#1 - \#0), the retrospective feed mechanism is improved \mname by an average of 0.58\% (\#2 - \#1), and the enhanced recurrence mechanism makes a large contribution of 2.55 percentage points on average (\#3 - \#2). By comparing \#3 with \#4, we see that segment-level recurrence is necessary for modeling long documents and produces a 4.92 percentage point improvement on average. Considering different types of tasks, we observe that on Hyperpartisan, an extremely long text classification dataset, a substantial improvement is achieved using the segment-reordering objective (1.5\% point). This indicates that the \texttt{[CLS]} token, pretrained using the segment-reordering objective, is more adaptable to the document-level text classification task. Moreover, we observed a stable performance gain across all tasks using the enhanced recurrence mechanism.

\subsection{Hyperparameters for Language Modeling}
In Tab.~\ref{tab:param_lm}, we present the detailed hyperparameters used for our experiments, which are the same as the hyperparameters employed in Transformer-XL ~\citep{transformer_xl}.

\begin{table}[!h]
\resizebox{0.49\textwidth}{!}{
\begin{tabular}{l|cc}
\hline \hline
\textbf{Hyperparameters} & \textbf{\begin{tabular}[c]{@{}c@{}}WikiText-103\\ Base\end{tabular}} & \textbf{\begin{tabular}[c]{@{}c@{}}WikiText-103\\ Large\end{tabular}} \\ \hline
Layers                   & 16                                                                   & 18                                                                    \\
Hidden size              & 410                                                                  & 1,024                                                                  \\
Attention heads          & 10                                                                   & 16                                                                    \\
Training sequence length & 150                                                                  & 384                                                                   \\
Training memory length   & 150                                                                  & 384                                                                   \\
Testing sequence  length & 64                                                                   & 128                                                                   \\
Testing sequence length  & 640                                                                  & 1,600                                                                  \\
Batch size               & 64                                                                   & 128                                                                   \\
Learning rate            & 2.5e-4                                                               & 2.5e-4                                                                \\
Warmup steps             & 0                                                                    & 16,000                                                                 \\
Training steps           & 200k                                                                 & 400k                                                                  \\ \hline
\end{tabular}
}
\caption{Hyperparameters used for WikiText-103.}
\label{tab:param_lm}
\vspace{-0.6cm}
\end{table}

\subsection{Hyperparameters for Pre-Training}
As shown in Tab.~\ref{tab: pretraing_params}, we present the detailed hyperparameters adopted to pretraining \mname on English text corpora and Chinese text corpora. For comparisons, we follow the same optimization hyperparameters of RoBERTa$_\text{BASE}$ or RoBERTa$_\text{LARGE}$ \citep{liu2019roberta} for base-size or large-size model in English domain. As for Chinese \mname, we follow the same optimization hyperparameters of ERNIE 2.0$_\text{BASE}$. 

\begin{table}[!h]
\resizebox{0.49\textwidth}{!}{
\begin{tabular}{l|ccc}
\hline \hline
\textbf{Hyperparameters} & \multicolumn{2}{c}{English}           & Chinese              \\ \cline{2-4} 
                         & \textbf{BASE}        & \textbf{LARGE} & \textbf{BASE}        \\ \hline
Layers                   & 12                   & 24             & 12                   \\
Hidden size              & 768                  & 1,024           & 768                  \\
Attention heads          & 12                   & 16             & 12                   \\
Training steps           & 400K                 & 100K           & 300K                 \\
Batch size               & 2,560                 & 3,920           & 2,560                 \\
Learning rate            & 1e-4                 & 1e-4           & 1e-4                 \\
Warmup steps             & 4,000                 & 4,000           & 4,000                 \\
Adam (beta1,beta2)                    &  (0.9, 0.999)         & (0.9, 0.999)     &  (0.9, 0.999)         \\
Adam (epsilon)                     &  1e-6                    & 1e-6           &  1e-6                    \\
Learning rate schedule   &  Linear                     & Linear         &  Linear                     \\
Weight decay             & 0.01 & 0.01           & 0.01 \\
Dropout                  & 0.1 & 0.1            & 0 \\
GPU (Nvidia V100)        & 40                   & 80             & 40                   \\ \hline
\end{tabular}
}
\caption{Hyperparameters used for \mname pretraining.}
\label{tab: pretraing_params}
\vspace{-0.5cm}
\end{table}

\subsection{Hyperparameters for Fine-Tuning}

\subsubsection{Long Text Classification tasks}
The finetuning hyperparameters for IMDB \citep{maas2011learning} and Hyperpartisan \citep{kiesel2019semeval} are presented in Tab.~\ref{tab: params_class_en}.

\begin{table}[!h]
\centering
\resizebox{0.45\textwidth}{!}{
\begin{tabular}{l|cc|cl}
\hline \hline
\textbf{Hyperparameters} & \multicolumn{2}{c|}{BASE}              & \multicolumn{2}{c}{LARGE}              \\ \cline{2-5} 
                         & \textbf{IMDB} & \textbf{HYP} & \textbf{IMDB} & \textbf{HYP} \\ \hline
Batch size               & 32              & 32                       & 32              & 16                       \\
Learning rate            & 7e-5              & 1e-4                       & 1e-5              & 4e-6                       \\
Epochs                   & 3              & 15                       & 3              & 15                       \\
LR schedule              & linear              & linear                       & linear              & linear                       \\
Layerwise LR decay       & 1              & 0.7                       & 0.9              & 1                        \\
Warmup proportion        & 0.1              &  0.1                      & 0.1              &  0.1                      \\
Weight decay             & 0.01              & 0.01                       & 0.01              &  0.01                      \\ \hline
\end{tabular}}
\caption{Hyperparameters used for finetuning on IMDB and Hyperpartisan (HYP).}
\label{tab: params_class_en}
\vspace{-0.5cm}
\end{table}

\subsubsection{Document-level Question answering tasks}
The finetuning hyperparameters for TriviaQA \citep{welbl2018constructing} and HotpotQA \citep{yang2018hotpotqa} are presented in Tab.~\ref{tab:params_qa_en}. HQA-sent. is the model for coarse-grained evidence prediction, and we choose the evidence with the probability larger than a pre-defined threshold 1e-3 and 1e-5 for base and large models, respectively. HQA-span. is the model for span prediction.

\begin{table}[]
\centering
\resizebox{0.48\textwidth}{!}{
\begin{tabular}{l|ccc|ccc}
\hline \hline
\multirow{2}{*}{\textbf{Hyper.}} & \multicolumn{3}{c|}{BASE}                           & \multicolumn{3}{c}{LARGE}                         \\ \cline{2-7} 
                        & \textbf{TQA}    & \textbf{HQA-sent.} & \multicolumn{1}{l|}{\textbf{HQA-span.}} & \textbf{TQA}    & \textbf{HQA-sent.} & \multicolumn{1}{l}{\textbf{HQA-span.}} \\ \hline
Batch size              & 64     & 128       & 128                            & 64     & 32        & 32                            \\
Learning rate           & 3e-5   & 3e-5      & 1.5e-4                         & 5e-6   & 5e-6      & 1.5e-5                        \\
Epochs                  & 5      & 6         & 6                              & 3      & 5         & 5                             \\
LR schedule             & linear & linear    & linear                         & linear & linear    & linear                        \\
Layer-decay             & 0.8    & 1         & 0.8                            & 0.9    & 0.9       & 0.9                           \\
Warmup prop.            & 0.1    & 0.1       & 0.1                            & 0.1    & 0.1       & 0.1                           \\
Weight decay            & 0.01   & 0.01      & 0.01                           & 0.01   & 0.01      & 0.01                          \\ \hline
\end{tabular}}\caption{Finetuning hyperparameters on the TQA and HQA for base- and large-size \mname.}
\label{tab:params_qa_en}
\end{table}

\subsubsection{Keyphrase Extraction task}
The finetuning hyperparameters for the OpenKP \citep{xiong2019open} dataset are presented in Tab.~\ref{tab:params_kp_en}. 

\begin{table}[!h]
\centering
\resizebox{0.27\textwidth}{!}{
\begin{tabular}{l|c}
\hline \hline
\textbf{Hyperparameters}  & \textbf{OpenKP} \\ \hline
Batch size                                &  32             \\
Learning rate                              & 1.5e-4               \\
Epochs                                       & 5              \\
LR schedule                                & linear                 \\
Layerwise LR decay                         & 0.8               \\
Warmup proportion                          & 0.1                \\
Weight decay                             & 0.01                 \\ \hline
\end{tabular}
}
\caption{Finetuning hyperparameters on the OpenKP for base-size \mname.}
\label{tab:params_kp_en}
\vspace{-0.5cm}
\end{table}

\subsubsection{Chinese NLU tasks}
Tab.~\ref{tab:params_nlu_zh} lists the finetuning hyperparameters for Chinese NLU tasks including IFLYTEK \citep{iflytek}, THUCNews \citep{thucnews}, CMRC2018 \citep{cmrc2018}, DRCD \citep{drcd}, DuReader \citet{dureader}, C$^3$ \citep{c3} and CAIL2019-SCM \citep{cail}.

\begin{table}[!h]
\centering
\resizebox{0.47\textwidth}{!}{
\begin{tabular}{l|cccc}
\hline \hline
\textbf{Tasks} & \textbf{\begin{tabular}[c]{@{}c@{}}Batch \\ size\end{tabular}} & \textbf{\begin{tabular}[c]{@{}c@{}}Learning\\ rate\end{tabular}} & \textbf{Epochs} & \multicolumn{1}{l}{\textbf{Dropout}} \\ \hline
DRCD           & 64                                                             & 2.25-4                                                           & 5               & 0.1                                  \\
CMRC2018       & 64                                                             & 1.75e-4                                                          & 5               & 0.2                                  \\
DuReader       & 64                                                             & 2.75e-4                                                          & 5               & 0.1                                  \\
C3             & 24                                                             & 1e-4                                                              & 8               & 0.1                                  \\
CAIL           & 48                                                             & 5e-5                                                             & 15              & 0.1                                  \\
THU            & 16                                                             & 1.5e-4                                                           & 16              & 0.1                                  \\
IFK            & 16                                                             & 1.5e-4                                                           & 5               & 0.1                                  \\ \hline
\end{tabular}
}
\caption{Hyperparameters used for finetuning on Chinese NLU tasks. Note that the warmup proportion are set to 0.1 and the layerwise learning rate decay rate are set to 0.8 for all tasks.}
\label{tab:params_nlu_zh}
\vspace{-0.5cm}
\end{table}

\section{Attention Complexity}
Given a long document with length $L$, Longformer and BigBird usually applies a local attention with a window size of 512 tokens on the entire input resulting in $L*512$ token-to-token calculations. While the long document is fed twice as input and each input is sliced with a sliding window size of 512 tokens in \mname, which resulting in $2*\frac{L}{512}*512*(512+m)$ token-to-token calculations where $m$ is the memory length. Since $512 \ll L \text{ and } m \ll L$, the attention complexity of \mname is comparable to Longformer and BigBird which scales linearly with respect to the input length $L$, i.e., $O(L)$. Notably, the segments produced from the long document are fed one by one in \mname, leading to the lower spatial complexity.

\end{document}